\documentclass[sigconf]{acmart}

\usepackage{graphicx}
\usepackage{subfigure}
\usepackage{makecell}
\usepackage{lipsum}

\newcommand*{\affaddr}[1]{#1} 
\newcommand*{\affmark}[1][*]{\textsuperscript{#1}}

\AtBeginDocument{%
	\providecommand\BibTeX{{%
			\normalfont B\kern-0.5em{\scshape i\kern-0.25em b}\kern-0.8em\TeX}}}

\setcopyright{none}
\renewcommand\footnotetextcopyrightpermission[1]{}
\begin{document}

\title{
		Winograd Convolution: A Perspective from Fault Tolerance
	}

   \author{%
   Xinghua Xue\affmark[1], Haitong Huang\affmark[1], Cheng Liu\affmark[1], Ying Wang\affmark[1], Tao Luo\affmark[2], Lei Zhang\affmark[1]\\
   \affaddr{\affmark[1]Institute of Computing Technology, Chinese Academy of Sciences,\\ 
   University of Chinese Academy of Sciences, Beijing, China}\\
   \affaddr{\affmark[2]Institute of High Performance Computing, A*STAR}\\
   \{xuexinghua, liucheng, wangying2009, zlei\}@ict.ac.cn, \\
   huanghaitong21@mails.ucas.ac.cn,   luo\_tao@ihpc.a-star.edu.sg
   }

    \begin{abstract}
	Winograd convolution is originally proposed to reduce the computing overhead by converting multiplication in neural network (NN) with addition via linear transformation. Other than the computing efficiency, we observe its great potential in improving NN fault tolerance and evaluate its fault tolerance comprehensively for the first time. Then, we explore the use of fault tolerance of winograd convolution for either fault-tolerant or energy-efficient NN processing. According to our experiments, winograd convolution can be utilized to reduce fault-tolerant design overhead by 27.49\% or energy consumption by 7.19\% without any accuracy loss compared to that without being aware of the fault tolerance.\par
	\end{abstract}
	\maketitle	
\pagestyle{plain}
	
	

\section{Introduction}
Deep neural network (DNN) has been demonstrated to be successful in numerous domains of applications such as computer vision, natural language processing, and data analytics \cite{liu2017survey}. It attracts worldwide attentions and substantial research efforts have been devoted to improving DNNs from the perspective of precision, performance, energy efficiency, and so on \cite{deng2020model}. Fault tolerance is also a vital DNN metric yet has generally been overlooked \cite{reagen2018ares}. 

Although DNNs are inherently fault-tolerant to errors or noise \cite{torres2017fault} \cite{temam2012defect} \cite{reagen2016minerva} \cite{deng2015retraining}, improving the fault tolerance of DNNs remains highly demanded as it can potentially bring in multi-folded benefits. First of all, the underlying computing engines fabricated on silicon are not perfect and can suffer increasing soft errors because of the continuously lower power supply and higher transistor density over the years \cite{bohr2017cmos} \cite{Pandey2019GreenTPU}. The influence of soft errors on the DNN execution is almost unavoidable. They can cause dramatic accuracy drop and even catastrophic consequences for safety-critical DNN applications such as autonomous driving and medical diagnosis \cite{R2F2021TVLSI}. Although soft errors can be mitigated with fault-tolerant hardware designs, fault tolerant DNN models can greatly alleviate or even completely avoid the fault-tolerant hardware design overhead \cite{FTT-NAS2020ASPDAC} \cite{he2018axtrain}. Second, when DNNs are applied to applications like entertainment that are insensitive to the soft errors, the fault tolerance of DNNs can also be explored to relax the requirements of absolute correctness of the DNN execution, which can improve the energy efficiency substantially with techniques such as aggressive voltage scaling\cite{reagen2016minerva} \cite{tu2018rana}, overclocking \cite{deng2015retraining} \cite{Li2019squeezing}, and model pruning \cite{deng2020model}. Thereby, improving the fault tolerance of DNNs is beneficial in general and required for safety-critical applications particularly.

A few work have been proposed to improve the fault tolerance of DNNs from different angles. The authors in \cite{jia2018calibrating} \cite{R2F2021TVLSI} opted to mitigate soft errors in neural network accelerators with in-situ model retraining. Despite the fault tolerance to the targeted fault configurations, they may require additional retraining to adapt to a different fault scenario. The authors in \cite{FTT-NAS2020ASPDAC} \cite{ning2021ftt} utilized network architecture search (NAS) to search both high-precision and fault-tolerant neural networks. Basically, they have fault tolerance and precision combined into a unified metric for the network candidate selection during NAS. Hence, the network design space shrinks because of the more stringent design constraints, and the resulting DNN models usually get more complex and induce additional overhead. According to their experiments, the searched neural network models have quite some duplicated cells. Jack Kosaian et al. \cite{kosaian2021arithmetic} proposed a lightweight algorithm-based fault tolerance (ABFT) approach for error mitigation of neural network processing on GPUs. Elbruz Ozen et al. \cite{Ozen2019sanity-check} proposed Sanity-Check, which employs spatial and temporal checksums to protect fully-connected and convolutional layers in deep neural networks. Although these approaches improve the DNN fault tolerance, they generally require additional computing overhead or compromise the model generality.

Unlike prior work, we aim to enhance the inherent fault tolerance of DNNs without compromising other features of the models. With the observation that bit flip errors in input operands of multiplication typically can cause more severe computing errors over the golden outputs and multiplication is generally more sensitive to the soft errors compared to addition, we envision that winograd convolution which is originally proposed to reduce the computing overhead by reducing the number of multiplications \cite{lavin2016fast} is potentially beneficial to the fault tolerance of DNN execution. As winograd convolution offers an efficient way to conduct the DNN computing almost for free \cite{lu2017evaluating} \cite{liu2018efficient} \cite{shen2018towards} \cite{jia2018optimizing}, we investigate the fault tolerance of winograd-based DNN computing comprehensively in this work. In order to evaluate the fault tolerance of winograd-based DNNs, we develop an operation-level fault injection approach rather than reusing existing neuron-level fault injection methods \cite{mahmoud2020pytorchfi} \cite{chen2020tensorfi} \cite{reagen2018ares} which can hardly differentiate the standard convolution and winograd convolution due to the lack of fault simulation of details in each neuron computing. 

On top of the fault tolerance evaluation of winograd convolution, we further explore two different ways to take advantage of the fault tolerance brought by winograd convolution. One of them is to reduce DNN fault-tolerant design overhead. Specifically, we propose a fine-grained triple modular redundancy (TMR) protection approach and demonstrate the reduced DNN fault-tolerant design overhead under different fault configurations. The other one is to reduce DNN energy consumption. Basically, we can conduct more aggressive voltage scaling of computing engines while the timing errors induced by the voltage scaling can be tolerated by the winograd convolution. Hence, the energy consumption can be reduced without any model accuracy loss. 

The contributions of this work can be summarized as follows.\par
\begin{itemize}
\item We discover and investigate the fault tolerance of winograd-based DNNs comprehensively for the first time and explore the use of the enhanced fault tolerance of DNNs for either soft error mitigation or computing energy reduction.  

\item We propose a fine-grained TMR-based redundancy approach for DNNs to make best use of winograd convolution fault tolerance. According to our experiments, winograd convolution with being aware of the fault tolerance shows 61.21\% and 27.49\% less computing overhead compared to standard convolution and winograd convolution without being aware of the inherent fault tolerance respectively. 

\item We take advantage of the winograd convolution fault tolerance to enable more aggressive supply voltage scaling of a typical deep learning accelerator. According to our experiments, this approach achieves 42.89\% and 7.19\% energy reduction  compared to voltage scaling with standard convolution and winograd convolution without being aware of the inherent fault tolerance.
\end{itemize}

\section{Background and Related Work}





\subsection{Fault Tolerance of Neural Networks}
Soft errors that are usually caused by high-energy particles hitting on electronic devices can lead to malfunction and considerable prediction accuracy loss of neural networks that eventually rely on the silicon-based computing engines such as neural network accelerators, GPUs, and CPUs. Although conventional fault-tolerant design approaches such as triple modular redundancy, ECC, and algorithm-based fault tolerance (ABFT) \cite{kosaian2021arithmetic} \cite{Ozen2019sanity-check} can potentially alleviate the influence of soft errors, they usually require considerable computing overhead. In contrast, it is more flexible and cost-effective to take advantage of the inherent fault tolerance of the neural networks and alleviate the fault-tolerant design overhead at least. The authors in \cite{dutta2019codenet} \cite{zhao2017aep} \cite{hacene2019training} explored fault tolerance of neural networks with retraining. Prasenjit Dey et al. \cite{dey2017regularizing} proposed to penalize the system errors with regularizing terms to make MLP robust to various errors such as link failures, multiplicative noise, and additive noise. Elbruz Ozen et al. \cite{ozen2021snr} also proposed a novel regularization term to manipulate the parameter distributions at each layer and tighten error margins. The resulting models show improved resilience noticeably against soft errors in the neural network variables. Unlike the work that retain the model architectures, the authors in \cite{ning2021ftt} \cite{FTT-NAS2020ASPDAC} proposed to take fault tolerance into consideration in neural architecture search (NAS) framework and obtain a new network architecture that fulfills both the fault tolerance metric and accuracy metric at the same time. More fault-tolerant neural network design approaches can be found in recent surveys \cite{mittal2020survey} \cite{robust2020Mutlu} \cite{liu2017survey}. In summary, we can conclude that improving the fault tolerance of neural networks usually takes only minor or moderate overhead and thus is an effective way to mitigate soft errors in silicon especially when compared to expensive conventional fault-tolerant approaches.


Neural network fault tolerance can also be utilized to improve the performance or energy efficiency of neural network processing other than the soft error mitigation as demonstrated in prior work \cite{reagen2016minerva} \cite{tu2018rana} \cite{Li2019squeezing} \cite{deng2015retraining} \cite{liu2019exploiting} \cite{marty2020safe}. For instance, Fengbin Tu et al. \cite{tu2018rana} proposed to take eDRAM with relaxed retention as the data buffers of neural network accelerators for lower energy consumption while the data corruption induced by the aggressive data retention is tolerated by the neural network models. Thibaut Marty et al. \cite{marty2020safe} adopted timing speculation coupled with lightweight error detection as an approach to improve
the performance of hardware accelerators for CNNs. Brandon Reagen et al. \cite{reagen2016minerva} explored the inherent neural network fault tolerance to lower the neural network energy consumption by scaling the SRAM supply voltage. These approaches generally achieve significant energy saving or performance improvement by relaxing the requirements of absolute correctness of the neural network processing with only minor or no accuracy drop of the neural networks. These work demonstrate the great potential of inherent fault tolerance of the neural networks.


\subsection{Winograd Convolution}
Winograd convolution converts the matrix multiplication in standard convolution into element-wise multiplication by linearly transforming the input feature map and convolution kernels to a different domain of data representation. The element-wise multiplication results can be restored to the standard feature map domain with the corresponding inverse linear transformation. With the transform-calculation-inverse transformation process, the number of multiplication operations is considerably reduced and replaced with more cost-effective addition operations compared to standard convolution, which significantly enhances the computing efficiency. A basic two-dimensional winograd convolution is formulated in \autoref{eq:winograd-eq}. $Y$ denotes the output feature, $d$ denotes the input feature, $g$ denotes the filter, $G$ denotes the linear transformation matrix, $B$ denotes the input transformation matrix, and $A$ denotes the output transformation matrix. $G$, $B$, and $A$ are all constant matrices. When winograd convolution is conducted on small windows including 3×3 filter with unit stride, there will be no accuracy penalty. Even when the convolution filter and stride are larger, they can also be split to small ones according to the decomposable winograd method proposed in \cite{huang2020dwm}. In this case, winograd convolution can be processed without any accuracy penalty. Winograd convolution has been explored from various angles such as quantization \cite{li2020lance} \cite{fernandez2020searching}, tiling \cite{shen2019toward}, hardware acceleration \cite{lu2018spwa}, and \cite{lu2017evaluating} for efficient DNN computing. As far as we know, there is still a lack of investigation of winograd convolution from the perspective of fault tolerance. 
\vspace{-1mm}
\begin{equation}
Y =  A^{T}[[Gg_{k,c}G^{T} ] \odot[B^{T} d_{c,b}B]]A\label{eq:winograd-eq}
\end{equation}

In this work, we aim to investigate the fault tolerance of winograd-based convolution neural networks and take advantage of the improved neural network fault tolerance for either soft error mitigation or energy saving. 

\section{Fault Tolerance of Winograd DNNs}
\subsection{Fault Injection Platform}
Winograd convolution is conducted in layer wise and does not change the value of the neurons in DNNs with lossless conversion. As a result, neural network fault injection platforms such as TensorFI \cite{chen2020tensorfi} and PyTorchFI \cite{mahmoud2020pytorchfi} that generally have bit errors injected to neurons and weights can not differentiate the influence of soft errors on neural networks processed with standard convolution and winograd convolution. FIdelity \cite{he2020fidelity} has more hardware details considered, but it does not support winograd convolution yet. To that end, we propose an operation-level fault injection platform that has random soft errors injected to the results of primitive operations i.e. multiplication and addition in the neural network processing. It is implemented on top of PyTorch and open sourced on github. It focuses on computing of neural network processing rather than a specific computing engine. 

\begin{figure}
\begin{center}
\vspace{-10mm}
\includegraphics[scale=1.1]{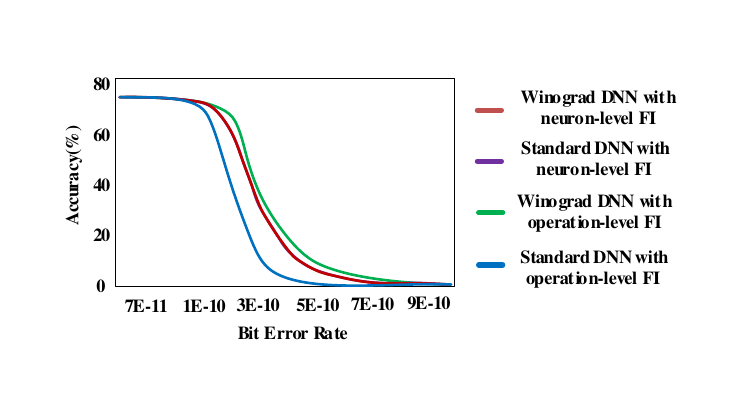}
\end{center}
\vspace{-11mm}
\caption{Comparison of the neuron-level fault injection (FI) and the operation-level FI.}
\vspace{-5mm}
\label{diff_analy_meth}
\end{figure}

To illustrate the precision of the proposed operation-level fault injection platform, we take VGG19 quantized with 16bit fixed point on CIFAR-100 dataset as an example and evaluate the model accuracy when different number of bit flip errors are injected. Bit error rate that denotes the probability of a bit flip in an operation is utilized as the soft error metric. We have the model accuracy of VGG19 with both standard convolution and winograd convolution evaluated on both neuron-level fault injection platform and operation-level fault injection platform when the bit error rate ranges from 7E-11 to 9E-10. The experiment result is shown in \autoref{diff_analy_meth}. It can be seen that there is no accuracy difference between standard convolution and winograd convolution under different bit error rate while the difference is clearly observed when the fine-grained operation-level fault injection platform is adopted. On the other hand, the general trend of the model accuracy under different bit error rate obtained from the two different fault injection platforms remains consistent in spite of the fault simulation accuracy difference. 

\subsection{Fault Tolerance Evaluation} \label{sec:evaluation}

With the operation-level fault injection platform, we investigate the fault tolerance of winograd DNNs from different angles. First of all, we evaluate the neural network accuracy calculated with standard convolution and winograd convolution respectively under distinct bit error rate and present an end-to-end fault tolerance comparison of the different computing methods of neural networks. Second, we conduct a layer-wise fault tolerance evaluation of the neural networks with standard convolution and winograd convolution respectively to investigate the winograd induced fault tolerance improvement across the different layers. Third, we evaluate the influence of soft errors on different types of operations in neural network processing with both standard convolution and winograd convolution. Finally, we conclude the fault tolerance comparison of the two different calculation methods. 

\subsubsection{Evaluation Setup}
In this evaluation, we have DenseNet169 on ImageNet, ResNet50 on ImageNet, VGG19 on CIFAR-100, and GoogleNet on CIFAR-10 utilized as the benchmark neural networks. Each neural network is quantized to a 8bit fixed point version and a 16bit fixed point version respectively. The bit error rate ranges from 0 to 1E-7 to cover the entire model accuracy drop process of the benchmark neural networks.

\subsubsection{Network-wise Fault Tolerance Evaluation}
\begin{figure}
\vspace{-5mm}
\begin{center}
\includegraphics[scale=0.52]{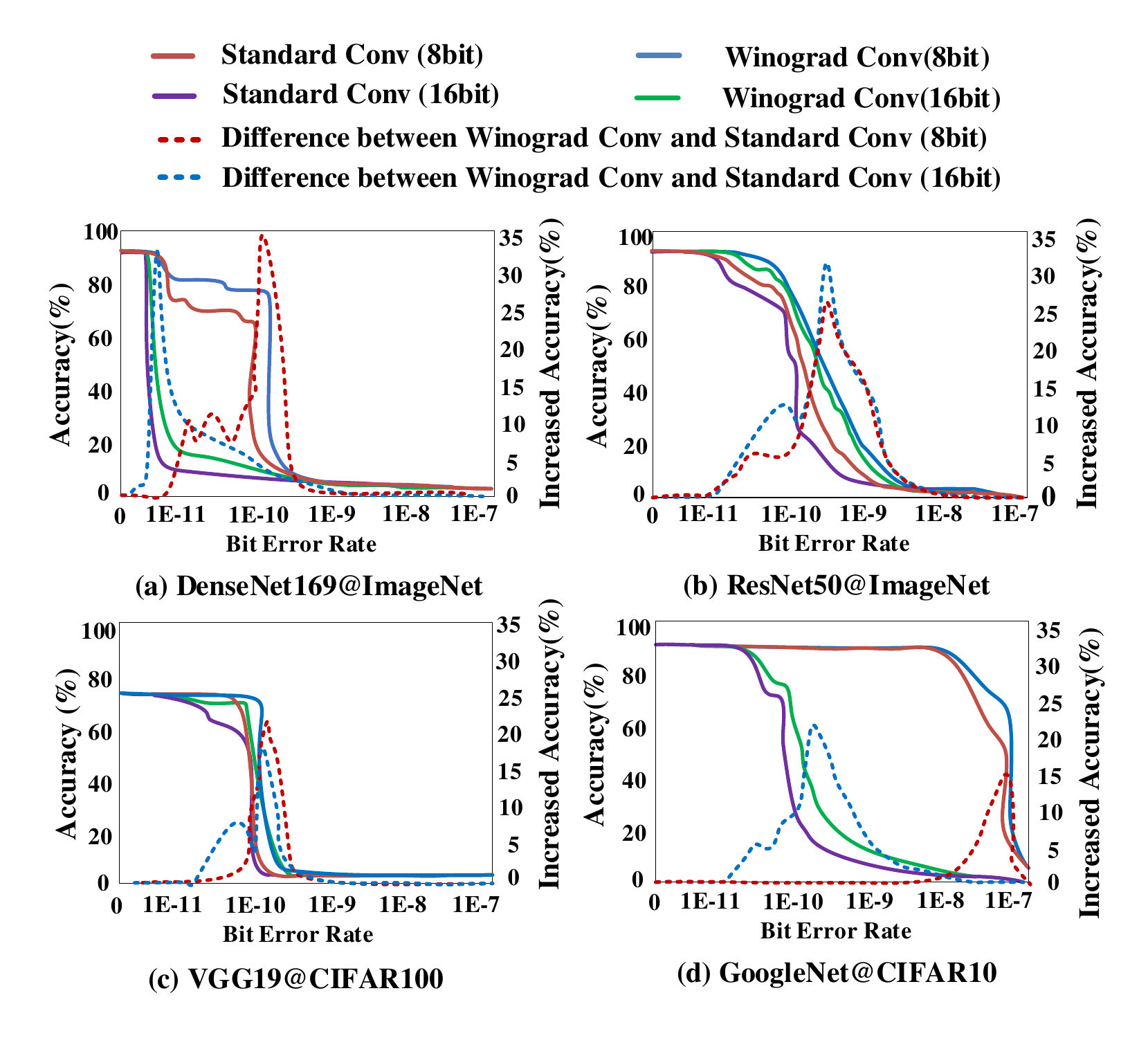}
\end{center}
\vspace{-5mm}
\caption{Accuracy of benchmark neural networks calculated with standard convolution and winograd convolution.}
\vspace{-6mm}
\label{fig:accuracy-result}
\end{figure}

We evaluated and compared the accuracy of the benchmark neural networks implemented with standard convolution and winograd convolution under different bit error rate. The experiment result is shown in \autoref{fig:accuracy-result}. It can be observed that neural network implemented with winograd convolution generally shows significant higher accuracy compared to that implemented with standard convolution. The accuracy improvement of winograd convolution relative to standard convolution as shown in dotted line in this figure reveals that the improvement can be up to 35\% and it holds true despite the network architectures, data width, and bit error rate. Nevertheless, the benefits can vary. For instance, models quantized with int16 are more vulnerable than that with int8 because bit filp for int16 can cause larger data variation on average. The network architectures also lead to dramatic difference. DenseNet and ResNet on top of the same dataset exhibit distinct accuracy improvement using winograd convolution. DenseNet shows rather sharp model accuracy improvement while ResNet shows more smooth model accuracy improvement. Hence, more fine-grained fault analysis is required to understand the fault tolerance of winograd DNNs comprehensively.


\subsubsection{Layer-wise Fault Tolerance Evaluation} \label{sec:layer-vulnerability}
To gain insight of the fault tolerance of winograd DNNs, we conduct layer-wise fault analysis of the neural networks with both winograd convolution and standard convolution, which is also an important basis for selective model protection. To characterize a layer fault tolerance of neural networks, we take VGG19 on CIFAR-100 as an example and conduct fault simulation of two different configurations. In one configuration, we have random bit errors injected to operations of the entire neural network. On the other configuration, we have random bit errors injected to operations of the entire neural networks except the layer under evaluation. The model accuracy obtained with the first fault simulation is denoted as baseline. The model accuracy difference of that obtained with the second fault simulation relative to the baseline represents the fault sensitivity of the layer under evaluation. Larger fault sensitivity indicates that higher model accuracy can be recovered from the baseline when the evaluated layer is fault-free. Hence, the corresponding layer is more critical to the fault tolerance of the entire neural network. The layer-wise model accuracy of the entire neural network implemented with both standard convolution and winograd convolution can be obtained with the same fault simulation approach. They are denoted as ST-Conv, WG-Conv respectively while the baselines are denoted as ST-Conv-Base and WG-Conv-Base accordingly. 

\begin{figure}
\begin{center}
\vspace{-8mm}
\includegraphics[width=0.48\textwidth]{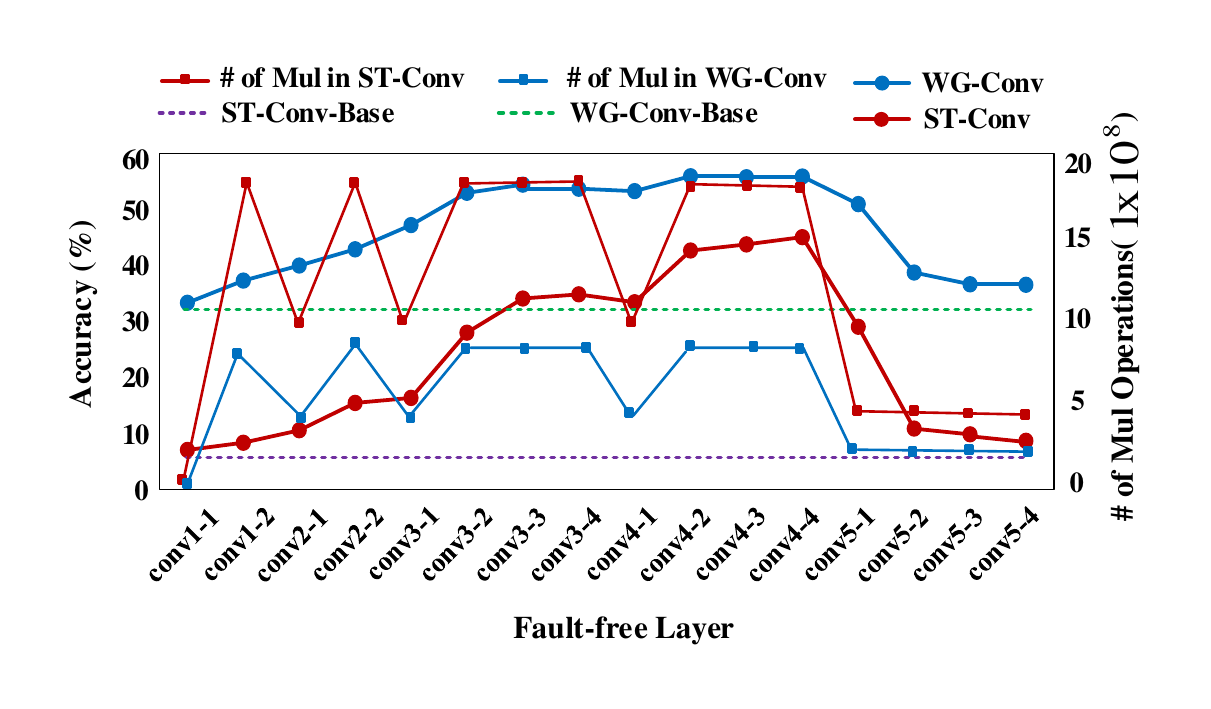}
\end{center}
\vspace{-8mm}
\caption{Accuracy of VGG19 on CIFAR-100 with one fault-free layer while the rest of layers are injected with bit error rate 3E-10. We have the neural network implemented with standard convolution and winograd convolution respectively. The base accuracy refers to the occasion when all the neural network layers are injected with the same bit error rate.}
\vspace{-6mm}
\label{fig:layer-accuracy}
\end{figure}

The experiment result is shown in \autoref{fig:layer-accuracy}. It reveals that the centering layers of the neural network are more sensitive to the bit errors while the layers in the beginning and the end of the network are less sensitive. It is probably because of the difference in operation number involved in each layer, as the layer-wise model accuracy is roughly consistent with the number of multiplication operations involved in each layer according to \autoref{fig:layer-accuracy}. Basically, the layers with more operations are likely to induce higher accuracy improvement when they are fault-free. In general, the layer-wise accuracy of the neural network implemented with winograd convolution is generally much higher than that implemented with standard convolution. The trend of the layer-wise accuracy is roughly consistent for neural networks implemented with both standard convolution and winograd convolution.  

\subsubsection{Operation Type Fault Tolerance Evaluation} \label{sec:type-vulnerability}
\begin{figure}
\begin{flushleft}
\vspace{-9mm}
\includegraphics[scale=0.8]{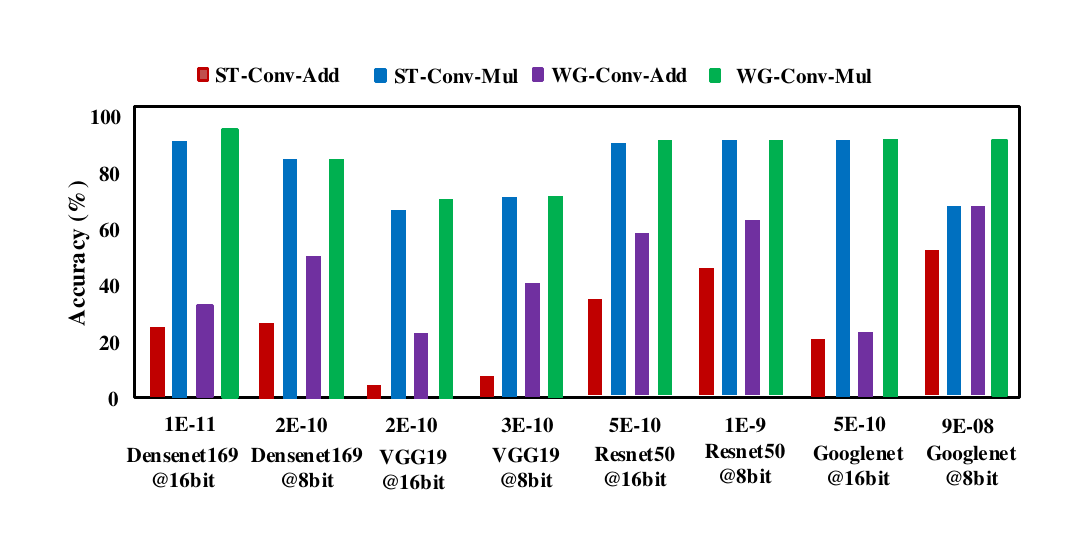}
\vspace{-10mm}
\end{flushleft}
\caption{Accuracy of neural networks with fault-free addition or fault-free multiplication under different bit error rate. Neural network implemented with standard convolution and winograd convolution are considered as well.}
\vspace{-5mm}
\label{fig:diff_ope}
\end{figure}

As winograd gre-atly changes the number of multiplication and addition in neural networks, we also investigate the fault tolerance from the perspective of operation types. The evaluation metric is similar to that of layer-wise accuracy. Assume that the entire model is exposed to random bit errors, then the model accuracy when the multiplication operations are kept fault-free can be used to measure the sensitivity of the multiplication operations to bit errors. Higher accuracy indicates that these operations are more vulnerable and need to be protected with higher priority. Similarly, the sensitivity of operations in different neural networks under different bit error rate can be obtained. The experiment result is shown in \autoref{fig:diff_ope}. Note that ST-Conv-Add and ST-Conv-Mul represent addition and multiplication in standard convolution respectively while WG-Conv-Add and WG-Conv-Mul represent addition and multiplication in winograd convolution accordingly. The experiment result shows that multiplication operations are more vulnerable than addition operations in all the different neural networks under various bit error rate. This also holds true for both standard convolution and winograd convolution. In addition, we notice that the accuracy of WG-Conv-Add with more addition included is generally higher than that of ST-Conv-Add but it remains much less important to the accuracy of the entire neural network. In contrast, the accuracy of WG-Conv-Mul with much less multiplications is still comparable to that of ST-Conv-Mul, which reveals that winograd convolution is more cost-effective for protection. 

\section{Exploring Winograd Fault Tolerance}
We already demonstrate significant fault tolerance improvement with winograd convolution and the internal difference in terms of layers and operation types in \autoref{sec:evaluation}. We further illustrate how the winograd convolution fault tolerance can be explored for more efficient fault tolerance and higher energy efficiency. 

\subsection{Using Winograd for Fault Tolerance}
TMR is a key strategy to mitigate soft errors in silicon. Unlike prior layer-wise TMR \cite{R2F2021TVLSI}, we propose a fine-grained heuristic TMR instead. The basic idea is to select the most vulnerable layer of a neural network with layer-wise vulnerability factor but protecting only a fraction of the operations in the layer rather than the entire layer. Note that the vulnerability factor of each layer is defined as the accuracy improvement when it is fault-free over when it is faulty. It can be obtained from the analysis in \autoref{sec:layer-vulnerability}. The operations to be selected is randomly chosen to ensure that the proposed TMR protection approach can be implemented efficiently on various computing engines. According to the analysis in \autoref{sec:type-vulnerability}, multiplication operations are much more vulnerable. Thus, we have multiplication operations selected with higher priority. The TMR protection is performed iteratively and a fraction of the operations will be protected in each iteration until the design goal i.e. accuracy is fulfilled. 

\begin{figure}
\begin{center}
\vspace{-8mm}
\includegraphics[scale=0.9]{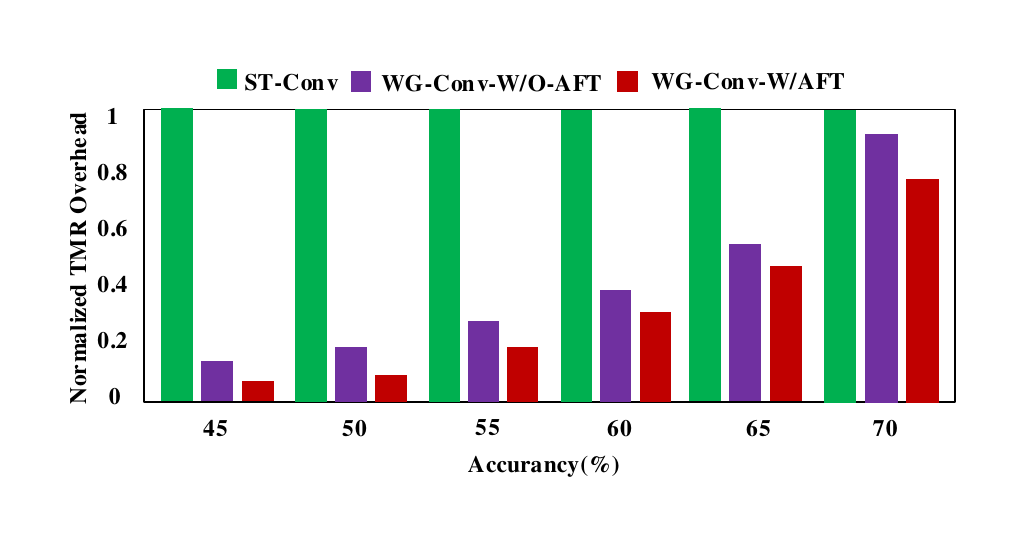}
\vspace{-9mm}
\end{center}
\caption{Normalized TMR overhead with different model accuracy requirements when bit error rate is 3E-10.}
\vspace{-9mm}
\label{fig:pro}
\end{figure}

We have the proposed TMR protection approach applied to three different neural network implementations including standard convolution (ST-Conv), winograd convolution without being aware of the fault tolerance (WG-Conv-W/O-AFT), and winograd convolution with being aware of the fault tolerance (WG-Conv-W/AFT). Note that WG-Conv-W/O-AFT evaluates the vulnerability factor of each layer with standard convolution and utilizes the same TMR protection option with ST-Conv because it is not aware of the fault tolerance of winograd convolution. The major difference between WG-Conv-W/O-AFT and ST-Conv is that WG-Conv-W/O-AFT conducts the neural network processing and protection on top of winograd convolution while ST-Conv conducts on standard convolution. In contrast, WG-Conv-W/AFT conducts both the layer-wise vulnerability factor analysis and TMR on top of winograd convolution. VGG19 quantized with int16 on CIFAR-100 is used as the benchmark example and its original model accuracy is 72.6\%. The bit error rate is set to be 3E-10. Since the TMR protection is essentially a design trade-off between TMR overhead and model accuracy. We set the model accuracy as the design goal and it ranges from 45\% to 70\% in this experiment. Then, we evaluate the TMR design overhead for the different design goals in this experiment. The TMR overhead is normalized to that of ST-Conv. The experiment result is presented in \autoref{fig:pro}. It can be observed that WG-Conv-W/O-AFT that greatly reduces the number of multiplication with winograd convolution achieves the same accuracy with much less TMR overhead. The benefit is mainly attributed to both the much less operations to be protected and part of the inherent fault tolerance of winograd convolution because of the lack of accurate layer vulnerability evaluation for the winograd convolution. In contrast, WG-Conv-W/AFT that fully explores the winograd convolution fault tolerance further reduces the TMR overhead by 27.49\% on average. 

\subsection{Using Winograd for Energy Efficiency}
\begin{figure}
\begin{center}
\vspace{-8mm}
\includegraphics[width=0.48\textwidth]{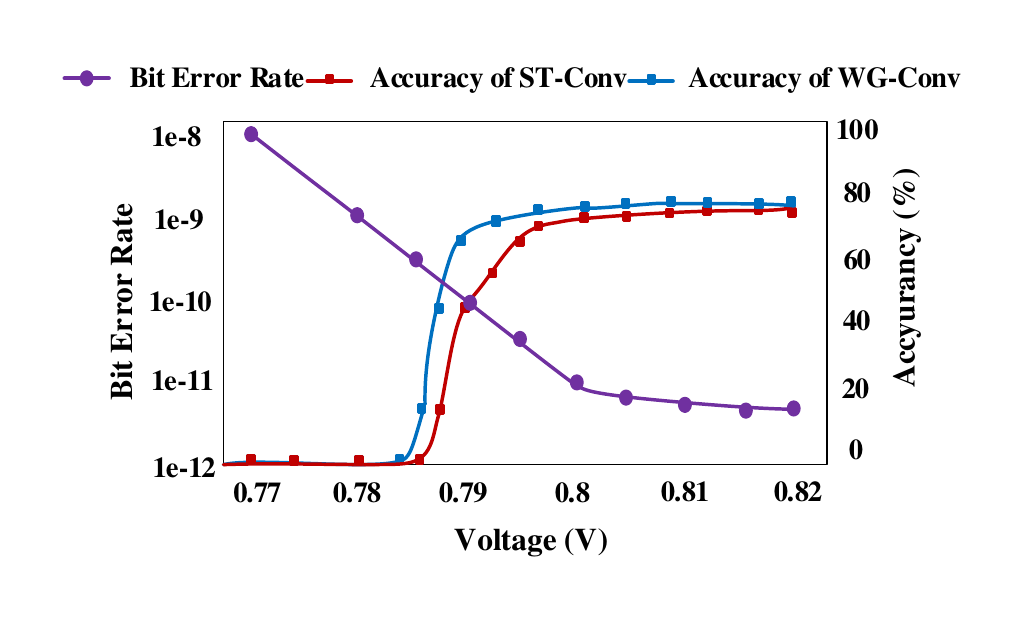}
\end{center}
\vspace{-8mm}
\caption{Bit error rate of the neural network accelerator and model accuracy of VGG19 on CIFAR-100 under different voltage scaling setups.}
\vspace{-4mm}
\label{fig:power2}
\end{figure}

We can also utilize neural network fault tolerance in combination with voltage scaling of the underlying computing engines for more energy-efficient processing \cite{du2020energy} \cite{whatmough2018dnn} \cite{rathore2020error} \cite{zhang2018thundervolt}. Basically, neural network fault tolerance relaxes the requirement of absolute computing of neural networks and we can have the computing engines for neural network processing working at lower voltage as long as the lower voltage induced timing errors and computing result variations can be tolerated by the neural networks. We take a typical neural network accelerator proposed in \cite{whatmough2018dnn} and VGG19 quantized with int16 for the experiment. The accelerator enables voltage scaling from 0.9V to 0.7V at 667MHz clock. The relationship between voltage and bit error rate as well as model accuracy is presented in \autoref{fig:power2}. Note that the voltage is constrained between 0.77V and 0.82V to highlight the major difference of interest in this experiment. We apply voltage scaling to three different implementations including standard convolution (ST-Conv), winograd convolution without being aware of the fault tolerance (WG-Conv-W/O-AFT), and winograd convolution with being aware of the fault tolerance (WG-Conv-W/AFT) under different model accuracy loss constraints ranging from 1\% to 10\%. Note that WG-Conv-W/O-AFT is a straightforward implementation of ST-Conv. The energy consumption depends on both power consumption of the accelerator according to \cite{whatmough2018dnn} and the runtime of the neural network processing which can be estimated with a simulator modified on top of Scale-Sim \cite{samajdar2018scale}. 

\begin{figure}
\begin{center}
\vspace{-5mm}
\includegraphics[width=0.48\textwidth]{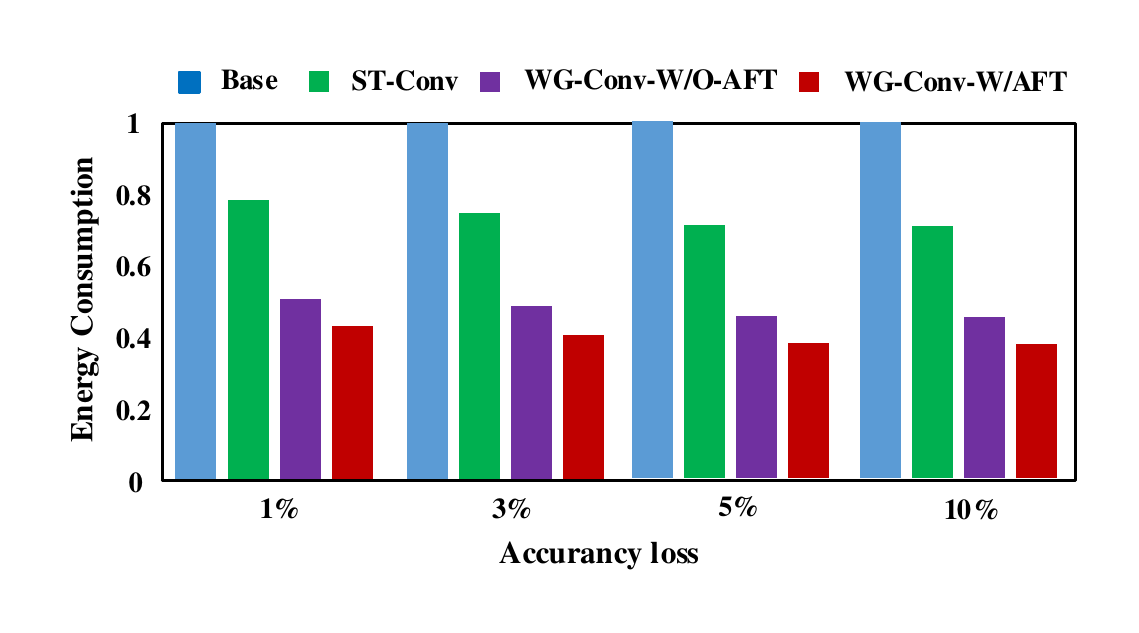}
\vspace{-10mm}
\end{center}
\caption{Voltage scaling assisted energy consumption of VGG19 quantized with int16 on CIFAR-100 implemented with ST-Conv, WG-Conv-W/O-AFT, and WG-Conv-W/-AFT under different accuracy loss constraints.}
\vspace{-8mm}
\label{power}
\end{figure}

The energy consumption of the three different implementations is further normalized to a standard convolution implementation without voltage scaling denoted as baseline which has the voltage set to be 0.9V. The comparison is shown in \autoref{power}. It can be seen that ST-Conv i.e. standard convolution taking advantage of its inherent fault tolerance with voltage scaling shows significant energy reduction compared to the baseline. We further explored the use of winograd convolution fault tolerance to improve the energy efficiency of the original DNN processing through voltage scaling. When we have ST-Conv implemented with winograd directly which is essentially WG-Conv-W/O-AFT, substantial energy reduction is obtained because of the reduced multiplication with winograd transformation, and reduce power consumption according to \cite{xygkis2018efficient}. Nevertheless, we can see that the energy consumption can be further reduced by 7.19\% on average according to the comparison between WG-Conv-W/O-AFT and WG-Conv-W/AFT. In summary, exploring the fault tolerance of winograd convolution enables further voltage scaling and additional energy reduction.

\section{Conclusion}
This paper discovers and studies the fault tolerance of winograd convolution that has been generally overlooked comprehensively for the first time. With this evaluation, we further explore the use of winograd fault tolerance for hardware-efficient fault-tolerant design and energy-efficient neural network processing respectively. For the hardware-efficient fault tolerance, we propose a novel fine-grained TMR protection approach which protects the neural network processing with layer-wise vulnerability analysis yet operation-wise TMR redundancy. The experiment reveals that winograd convolution fault tolerance can be utilized to reduce the TMR overhead by 61.21\% and 27.49\% on average compared to standard convolution and that without being aware of the winograd convolution fault tolerance respectively. For the energy-efficient neural network processing, we take advantage of the winograd convolution fault tolerance to enable more aggressive supply voltage scaling of a typical neural network accelerator. The experimental result demonstrates 42.89\% and 7.19\% energy reduction compared to standard convolution and that without being aware of winograd convolution fault tolerance respectively.


\bibliographystyle{ACM-Reference-Format}
\bibliography{ref}


\begin{thebibliography}{48}


\ifx \showCODEN    \undefined \def \showCODEN     #1{\unskip}     \fi
\ifx \showDOI      \undefined \def \showDOI       #1{#1}\fi
\ifx \showISBNx    \undefined \def \showISBNx     #1{\unskip}     \fi
\ifx \showISBNxiii \undefined \def \showISBNxiii  #1{\unskip}     \fi
\ifx \showISSN     \undefined \def \showISSN      #1{\unskip}     \fi
\ifx \showLCCN     \undefined \def \showLCCN      #1{\unskip}     \fi
\ifx \shownote     \undefined \def \shownote      #1{#1}          \fi
\ifx \showarticletitle \undefined \def \showarticletitle #1{#1}   \fi
\ifx \showURL      \undefined \def \showURL       {\relax}        \fi
\providecommand\bibfield[2]{#2}
\providecommand\bibinfo[2]{#2}
\providecommand\natexlab[1]{#1}
\providecommand\showeprint[2][]{arXiv:#2}

\bibitem[\protect\citeauthoryear{Alex}{Alex}{2019}]%
        {Ozen2019sanity-check}
\bibfield{author}{\bibinfo{person}{Ozen~Elbruz Alex}.}
  \bibinfo{year}{2019}\natexlab{}.
\newblock \showarticletitle{Sanity-Check: Boosting the Reliability of
  Safety-Critical Deep Neural Network Applications}. In
  \bibinfo{booktitle}{\emph{2019 IEEE 28th Asian Test Symposium (ATS)}}.
  \bibinfo{pages}{7--75}.
\newblock
\urldef\tempurl%
\url{https://doi.org/10.1109/ATS47505.2019.000-8}
\showDOI{\tempurl}


\bibitem[\protect\citeauthoryear{et~al.}{et~al.}{2011}]%
        {avirneni2011low}
\bibfield{author}{\bibinfo{person}{Avirneni~Naga et al.}}
  \bibinfo{year}{2011}\natexlab{}.
\newblock \showarticletitle{Low overhead soft error mitigation techniques for
  high-performance and aggressive designs}.
\newblock \bibinfo{journal}{\emph{IEEE Trans. Comput.}} \bibinfo{volume}{61},
  \bibinfo{number}{4} (\bibinfo{year}{2011}), \bibinfo{pages}{488--501}.
\newblock


\bibitem[\protect\citeauthoryear{et~al.}{et~al.}{2017a}]%
        {bohr2017cmos}
\bibfield{author}{\bibinfo{person}{Bohr Mark~T et al.}}
  \bibinfo{year}{2017}\natexlab{a}.
\newblock \showarticletitle{CMOS scaling trends and beyond}.
\newblock \bibinfo{journal}{\emph{IEEE Micro}} \bibinfo{volume}{37},
  \bibinfo{number}{6} (\bibinfo{year}{2017}), \bibinfo{pages}{20--29}.
\newblock


\bibitem[\protect\citeauthoryear{et~al.}{et~al.}{2020a}]%
        {chen2020tensorfi}
\bibfield{author}{\bibinfo{person}{Chen~Zitao et al.}}
  \bibinfo{year}{2020}\natexlab{a}.
\newblock \showarticletitle{TensorFI: A flexible fault injection framework for
  TensorFlow applications}. In \bibinfo{booktitle}{\emph{2020 IEEE 31st
  International Symposium on Software Reliability Engineering (ISSRE)}}. IEEE,
  \bibinfo{pages}{426--435}.
\newblock


\bibitem[\protect\citeauthoryear{et~al.}{et~al.}{2015a}]%
        {deng2015retraining}
\bibfield{author}{\bibinfo{person}{Deng~Jiacnao et al.}}
  \bibinfo{year}{2015}\natexlab{a}.
\newblock \showarticletitle{Retraining-based timing error mitigation for
  hardware neural networks}. In \bibinfo{booktitle}{\emph{2015 Design,
  Automation \& Test in Europe Conference \& Exhibition (DATE)}}. IEEE,
  \bibinfo{pages}{593--596}.
\newblock


\bibitem[\protect\citeauthoryear{et~al.}{et~al.}{2020b}]%
        {deng2020model}
\bibfield{author}{\bibinfo{person}{Deng~Lei et al.}}
  \bibinfo{year}{2020}\natexlab{b}.
\newblock \showarticletitle{Model compression and hardware acceleration for
  neural networks: A comprehensive survey}.
\newblock \bibinfo{journal}{\emph{Proc. IEEE}} \bibinfo{volume}{108},
  \bibinfo{number}{4} (\bibinfo{year}{2020}), \bibinfo{pages}{485--532}.
\newblock


\bibitem[\protect\citeauthoryear{et~al.}{et~al.}{2017b}]%
        {dey2017regularizing}
\bibfield{author}{\bibinfo{person}{Dey~Prasenjit et al.}}
  \bibinfo{year}{2017}\natexlab{b}.
\newblock \showarticletitle{Regularizing multilayer perceptron for robustness}.
\newblock \bibinfo{journal}{\emph{IEEE Transactions on Systems, Man, and
  Cybernetics: Systems}} \bibinfo{volume}{48}, \bibinfo{number}{8}
  (\bibinfo{year}{2017}), \bibinfo{pages}{1255--1266}.
\newblock


\bibitem[\protect\citeauthoryear{et~al.}{et~al.}{2019a}]%
        {dutta2019codenet}
\bibfield{author}{\bibinfo{person}{Dutta~Sanghamitra et al.}}
  \bibinfo{year}{2019}\natexlab{a}.
\newblock \showarticletitle{CodeNet: Training large scale neural networks in
  presence of soft-errors}.
\newblock \bibinfo{journal}{\emph{arXiv preprint arXiv:1903.01042}}
  (\bibinfo{year}{2019}).
\newblock


\bibitem[\protect\citeauthoryear{et~al.}{et~al.}{2020c}]%
        {du2020energy}
\bibfield{author}{\bibinfo{person}{Du~Yuxuan et al.}}
  \bibinfo{year}{2020}\natexlab{c}.
\newblock \showarticletitle{An Energy-Efficient Time-Domain Binary Neural
  Network Accelerator with Error-Detection in 28nm CMOS}. In
  \bibinfo{booktitle}{\emph{2020 IEEE Asia Pacific Conference on Circuits and
  Systems (APCCAS)}}. IEEE, \bibinfo{pages}{70--73}.
\newblock


\bibitem[\protect\citeauthoryear{et~al.}{et~al.}{2020d}]%
        {fernandez2020searching}
\bibfield{author}{\bibinfo{person}{Fernandez~Marques et al.}}
  \bibinfo{year}{2020}\natexlab{d}.
\newblock \showarticletitle{Searching for winograd-aware quantized networks}.
\newblock \bibinfo{journal}{\emph{arXiv preprint arXiv:2002.10711}}
  (\bibinfo{year}{2020}).
\newblock


\bibitem[\protect\citeauthoryear{et~al.}{et~al.}{2020e}]%
        {huang2020dwm}
\bibfield{author}{\bibinfo{person}{Huang~Di et al.}}
  \bibinfo{year}{2020}\natexlab{e}.
\newblock \showarticletitle{DWM: a decomposable winograd method for convolution
  acceleration}. In \bibinfo{booktitle}{\emph{Proceedings of the AAAI
  Conference on Artificial Intelligence}}, Vol.~\bibinfo{volume}{34}.
  \bibinfo{pages}{4174--4181}.
\newblock


\bibitem[\protect\citeauthoryear{et~al.}{et~al.}{2019b}]%
        {hacene2019training}
\bibfield{author}{\bibinfo{person}{Hacene~Ghouthi et al.}}
  \bibinfo{year}{2019}\natexlab{b}.
\newblock \showarticletitle{Training modern deep neural networks for
  memory-fault robustness}. In \bibinfo{booktitle}{\emph{2019 IEEE
  International Symposium on Circuits and Systems (ISCAS)}}. IEEE,
  \bibinfo{pages}{1--5}.
\newblock


\bibitem[\protect\citeauthoryear{et~al.}{et~al.}{2018a}]%
        {he2018axtrain}
\bibfield{author}{\bibinfo{person}{He~Xin et al.}}
  \bibinfo{year}{2018}\natexlab{a}.
\newblock \showarticletitle{AxTrain: Hardware-oriented neural network training
  for approximate inference}. In \bibinfo{booktitle}{\emph{Proceedings of the
  International Symposium on Low Power Electronics and Design}}.
  \bibinfo{pages}{1--6}.
\newblock


\bibitem[\protect\citeauthoryear{et~al.}{et~al.}{2020f}]%
        {he2020fidelity}
\bibfield{author}{\bibinfo{person}{He~Yi et al.}}
  \bibinfo{year}{2020}\natexlab{f}.
\newblock \showarticletitle{Fidelity: Efficient resilience analysis framework
  for deep learning accelerators}. In \bibinfo{booktitle}{\emph{2020 53rd
  Annual IEEE/ACM International Symposium on Microarchitecture (MICRO)}}. IEEE,
  \bibinfo{pages}{270--281}.
\newblock


\bibitem[\protect\citeauthoryear{et~al.}{et~al.}{2018b}]%
        {jia2018calibrating}
\bibfield{author}{\bibinfo{person}{Jia~Kaige et al.}}
  \bibinfo{year}{2018}\natexlab{b}.
\newblock \showarticletitle{Calibrating process variation at system level with
  in-situ low-precision transfer learning for analog neural network
  processors}. In \bibinfo{booktitle}{\emph{Proceedings of the 55th Annual
  Design Automation Conference}}. \bibinfo{pages}{1--6}.
\newblock


\bibitem[\protect\citeauthoryear{et~al.}{et~al.}{2018c}]%
        {jia2018optimizing}
\bibfield{author}{\bibinfo{person}{Jia~Zhen et al.}}
  \bibinfo{year}{2018}\natexlab{c}.
\newblock \showarticletitle{Optimizing N-dimensional, winograd-based
  convolution for manycore CPUs}. In \bibinfo{booktitle}{\emph{Proceedings of
  the 23rd ACM SIGPLAN Symposium on Principles and Practice of Parallel
  Programming}}. \bibinfo{pages}{109--123}.
\newblock


\bibitem[\protect\citeauthoryear{et~al.}{et~al.}{2021a}]%
        {kosaian2021arithmetic}
\bibfield{author}{\bibinfo{person}{Kosaian~Jack et al.}}
  \bibinfo{year}{2021}\natexlab{a}.
\newblock \showarticletitle{Arithmetic-Intensity-Guided Fault Tolerance for
  Neural Network Inference on GPUs}.
\newblock \bibinfo{journal}{\emph{arXiv preprint arXiv:2104.09455}}
  (\bibinfo{year}{2021}).
\newblock


\bibitem[\protect\citeauthoryear{et~al.}{et~al.}{2016a}]%
        {lavin2016fast}
\bibfield{author}{\bibinfo{person}{Lavin~Andrew et al.}}
  \bibinfo{year}{2016}\natexlab{a}.
\newblock \showarticletitle{Fast algorithms for convolutional neural networks}.
  In \bibinfo{booktitle}{\emph{Proceedings of the IEEE conference on computer
  vision and pattern recognition}}. \bibinfo{pages}{4013--4021}.
\newblock


\bibitem[\protect\citeauthoryear{et~al.}{et~al.}{2020g}]%
        {li2020lance}
\bibfield{author}{\bibinfo{person}{Li~Guangli et al.}}
  \bibinfo{year}{2020}\natexlab{g}.
\newblock \showarticletitle{Lance: efficient low-precision quantized winograd
  convolution for neural networks based on graphics processing units}. In
  \bibinfo{booktitle}{\emph{ICASSP 2020-2020 IEEE International Conference on
  Acoustics, Speech and Signal Processing (ICASSP)}}. IEEE,
  \bibinfo{pages}{3842--3846}.
\newblock


\bibitem[\protect\citeauthoryear{et~al.}{et~al.}{2017c}]%
        {lu2017evaluating}
\bibfield{author}{\bibinfo{person}{Lu~Liqiang et al.}}
  \bibinfo{year}{2017}\natexlab{c}.
\newblock \showarticletitle{Evaluating fast algorithms for convolutional neural
  networks on FPGAs}. In \bibinfo{booktitle}{\emph{2017 IEEE 25th Annual
  International Symposium on Field-Programmable Custom Computing Machines
  (FCCM)}}. IEEE, \bibinfo{pages}{101--108}.
\newblock


\bibitem[\protect\citeauthoryear{et~al.}{et~al.}{2018d}]%
        {lu2018spwa}
\bibfield{author}{\bibinfo{person}{Lu~Liqiang et al.}}
  \bibinfo{year}{2018}\natexlab{d}.
\newblock \showarticletitle{SpWA: An efficient sparse winograd convolutional
  neural networks accelerator on FPGAs}. In
  \bibinfo{booktitle}{\emph{Proceedings of the 55th Annual Design Automation
  Conference}}. \bibinfo{pages}{1--6}.
\newblock


\bibitem[\protect\citeauthoryear{et~al.}{et~al.}{2019c}]%
        {Li2019squeezing}
\bibfield{author}{\bibinfo{person}{L.~Li et al.}}
  \bibinfo{year}{2019}\natexlab{c}.
\newblock \showarticletitle{Squeezing the Last MHz for CNN Acceleration on
  FPGAs}. In \bibinfo{booktitle}{\emph{2019 IEEE International Test Conference
  in Asia (ITC-Asia)}}. \bibinfo{pages}{151--156}.
\newblock
\urldef\tempurl%
\url{https://doi.org/10.1109/ITC-Asia.2019.00039}
\showDOI{\tempurl}


\bibitem[\protect\citeauthoryear{et~al.}{et~al.}{2019d}]%
        {liu2019exploiting}
\bibfield{author}{\bibinfo{person}{Liu~Shanshan et al.}}
  \bibinfo{year}{2019}\natexlab{d}.
\newblock \showarticletitle{Exploiting asymmetry in eDRAM errors for
  redundancy-free error-tolerant design}.
\newblock \bibinfo{journal}{\emph{IEEE Transactions on Emerging Topics in
  Computing}} (\bibinfo{year}{2019}).
\newblock


\bibitem[\protect\citeauthoryear{et~al.}{et~al.}{2017d}]%
        {liu2017survey}
\bibfield{author}{\bibinfo{person}{Liu~Weibo et al.}}
  \bibinfo{year}{2017}\natexlab{d}.
\newblock \showarticletitle{A survey of deep neural network architectures and
  their applications}.
\newblock \bibinfo{journal}{\emph{Neurocomputing}}  \bibinfo{volume}{234}
  (\bibinfo{year}{2017}), \bibinfo{pages}{11--26}.
\newblock


\bibitem[\protect\citeauthoryear{et~al.}{et~al.}{2020h}]%
        {FTT-NAS2020ASPDAC}
\bibfield{author}{\bibinfo{person}{Li~Wenshuo et al.}}
  \bibinfo{year}{2020}\natexlab{h}.
\newblock \showarticletitle{FTT-NAS: Discovering Fault-Tolerant Neural
  Architecture}. In \bibinfo{booktitle}{\emph{2020 25th Asia and South Pacific
  Design Automation Conference (ASP-DAC)}}. \bibinfo{pages}{211--216}.
\newblock
\urldef\tempurl%
\url{https://doi.org/10.1109/ASP-DAC47756.2020.9045324}
\showDOI{\tempurl}


\bibitem[\protect\citeauthoryear{et~al.}{et~al.}{2018e}]%
        {liu2018efficient}
\bibfield{author}{\bibinfo{person}{Liu~Xingyu et al.}}
  \bibinfo{year}{2018}\natexlab{e}.
\newblock \showarticletitle{Efficient sparse-winograd convolutional neural
  networks}.
\newblock \bibinfo{journal}{\emph{arXiv preprint arXiv:1802.06367}}
  (\bibinfo{year}{2018}).
\newblock


\bibitem[\protect\citeauthoryear{et~al.}{et~al.}{2020i}]%
        {mahmoud2020pytorchfi}
\bibfield{author}{\bibinfo{person}{Mahmoud~Abdulrahman et al.}}
  \bibinfo{year}{2020}\natexlab{i}.
\newblock \showarticletitle{Pytorchfi: A runtime perturbation tool for dnns}.
  In \bibinfo{booktitle}{\emph{2020 50th Annual IEEE/IFIP International
  Conference on Dependable Systems and Networks Workshops (DSN-W)}}. IEEE,
  \bibinfo{pages}{25--31}.
\newblock


\bibitem[\protect\citeauthoryear{et~al.}{et~al.}{2020j}]%
        {marty2020safe}
\bibfield{author}{\bibinfo{person}{Marty~Thibaut et al.}}
  \bibinfo{year}{2020}\natexlab{j}.
\newblock \showarticletitle{Safe Overclocking for CNN Accelerators Through
  Algorithm-Level Error Detection}.
\newblock \bibinfo{journal}{\emph{IEEE Transactions on Computer-Aided Design of
  Integrated Circuits and Systems}} \bibinfo{volume}{39}, \bibinfo{number}{12}
  (\bibinfo{year}{2020}), \bibinfo{pages}{4777--4790}.
\newblock


\bibitem[\protect\citeauthoryear{et~al.}{et~al.}{2021b}]%
        {ning2021ftt}
\bibfield{author}{\bibinfo{person}{Ning~Xuefei et al.}}
  \bibinfo{year}{2021}\natexlab{b}.
\newblock \showarticletitle{FTT-NAS: Discovering fault-tolerant convolutional
  neural architecture}.
\newblock \bibinfo{journal}{\emph{ACM Transactions on Design Automation of
  Electronic Systems (TODAES)}} \bibinfo{volume}{26}, \bibinfo{number}{6}
  (\bibinfo{year}{2021}), \bibinfo{pages}{1--24}.
\newblock


\bibitem[\protect\citeauthoryear{et~al.}{et~al.}{2021c}]%
        {ozen2021snr}
\bibfield{author}{\bibinfo{person}{Ozen~Elbruz et al.}}
  \bibinfo{year}{2021}\natexlab{c}.
\newblock \showarticletitle{SNR: Squeezing Numerical Range Defuses Bit Error
  Vulnerability Surface in Deep Neural Networks}.
\newblock \bibinfo{journal}{\emph{ACM Transactions on Embedded Computing
  Systems (TECS)}} \bibinfo{volume}{20}, \bibinfo{number}{5s}
  (\bibinfo{year}{2021}), \bibinfo{pages}{1--25}.
\newblock


\bibitem[\protect\citeauthoryear{et~al.}{et~al.}{2019e}]%
        {Pandey2019GreenTPU}
\bibfield{author}{\bibinfo{person}{Pandey~Pramesh et al.}}
  \bibinfo{year}{2019}\natexlab{e}.
\newblock \showarticletitle{GreenTPU: Improving Timing Error Resilience of a
  Near-Threshold Tensor Processing Unit}. In \bibinfo{booktitle}{\emph{2019
  56th ACM/IEEE Design Automation Conference (DAC)}}. \bibinfo{pages}{1--6}.
\newblock


\bibitem[\protect\citeauthoryear{et~al.}{et~al.}{2016b}]%
        {reagen2016minerva}
\bibfield{author}{\bibinfo{person}{Reagen~Brandon et al.}}
  \bibinfo{year}{2016}\natexlab{b}.
\newblock \showarticletitle{Minerva: Enabling low-power, highly-accurate deep
  neural network accelerators}. In \bibinfo{booktitle}{\emph{2016 ACM/IEEE 43rd
  Annual International Symposium on Computer Architecture (ISCA)}}. IEEE,
  \bibinfo{pages}{267--278}.
\newblock


\bibitem[\protect\citeauthoryear{et~al.}{et~al.}{2018f}]%
        {reagen2018ares}
\bibfield{author}{\bibinfo{person}{Reagen~Brandon et al.}}
  \bibinfo{year}{2018}\natexlab{f}.
\newblock \showarticletitle{Ares: A framework for quantifying the resilience of
  deep neural networks}. In \bibinfo{booktitle}{\emph{2018 55th ACM/ESDA/IEEE
  Design Automation Conference (DAC)}}. IEEE, \bibinfo{pages}{1--6}.
\newblock


\bibitem[\protect\citeauthoryear{et~al.}{et~al.}{2020k}]%
        {rathore2020error}
\bibfield{author}{\bibinfo{person}{Rathore~Mallika et al.}}
  \bibinfo{year}{2020}\natexlab{k}.
\newblock \showarticletitle{Error Probability Models for Voltage-Scaled
  Multiply-Accumulate Units}.
\newblock \bibinfo{journal}{\emph{IEEE Transactions on Very Large Scale
  Integration (VLSI) Systems}} \bibinfo{volume}{28}, \bibinfo{number}{7}
  (\bibinfo{year}{2020}), \bibinfo{pages}{1665--1675}.
\newblock


\bibitem[\protect\citeauthoryear{et~al.}{et~al.}{2018g}]%
        {samajdar2018scale}
\bibfield{author}{\bibinfo{person}{Samajdar~Ananda et al.}}
  \bibinfo{year}{2018}\natexlab{g}.
\newblock \showarticletitle{Scale-sim: Systolic cnn accelerator simulator}.
\newblock \bibinfo{journal}{\emph{arXiv preprint arXiv:1811.02883}}
  (\bibinfo{year}{2018}).
\newblock


\bibitem[\protect\citeauthoryear{et~al.}{et~al.}{2018h}]%
        {shen2018towards}
\bibfield{author}{\bibinfo{person}{Shen~Junzhong et al.}}
  \bibinfo{year}{2018}\natexlab{h}.
\newblock \showarticletitle{Towards a uniform template-based architecture for
  accelerating 2D and 3D CNNs on FPGA}. In
  \bibinfo{booktitle}{\emph{Proceedings of the 2018 ACM/SIGDA International
  Symposium on Field-Programmable Gate Arrays}}. \bibinfo{pages}{97--106}.
\newblock


\bibitem[\protect\citeauthoryear{et~al.}{et~al.}{2019f}]%
        {shen2019toward}
\bibfield{author}{\bibinfo{person}{Shen~Junzhong et al.}}
  \bibinfo{year}{2019}\natexlab{f}.
\newblock \showarticletitle{Toward an efficient deep pipelined template-based
  architecture for accelerating the entire 2-D and 3-D CNNs on FPGA}.
\newblock \bibinfo{journal}{\emph{IEEE Transactions on Computer-Aided Design of
  Integrated Circuits and Systems}} \bibinfo{volume}{39}, \bibinfo{number}{7}
  (\bibinfo{year}{2019}), \bibinfo{pages}{1442--1455}.
\newblock


\bibitem[\protect\citeauthoryear{et~al.}{et~al.}{2020l}]%
        {robust2020Mutlu}
\bibfield{author}{\bibinfo{person}{Shafique~Muhammad et al.}}
  \bibinfo{year}{2020}\natexlab{l}.
\newblock \showarticletitle{Robust Machine Learning Systems: Challenges,Current
  Trends, Perspectives, and the Road Ahead}.
\newblock \bibinfo{journal}{\emph{IEEE Design Test}} \bibinfo{volume}{37},
  \bibinfo{number}{2} (\bibinfo{year}{2020}), \bibinfo{pages}{30--57}.
\newblock
\urldef\tempurl%
\url{https://doi.org/10.1109/MDAT.2020.2971217}
\showDOI{\tempurl}


\bibitem[\protect\citeauthoryear{et~al.}{et~al.}{2017e}]%
        {torres2017fault}
\bibfield{author}{\bibinfo{person}{Torres-Huitzil~Cesar et al.}}
  \bibinfo{year}{2017}\natexlab{e}.
\newblock \showarticletitle{Fault and error tolerance in neural networks: A
  review}.
\newblock \bibinfo{journal}{\emph{IEEE Access}}  \bibinfo{volume}{5}
  (\bibinfo{year}{2017}), \bibinfo{pages}{17322--17341}.
\newblock


\bibitem[\protect\citeauthoryear{et~al.}{et~al.}{2015b}]%
        {tan2015investigating}
\bibfield{author}{\bibinfo{person}{Tan~Li et al.}}
  \bibinfo{year}{2015}\natexlab{b}.
\newblock \showarticletitle{Investigating the interplay between energy
  efficiency and resilience in high performance computing}. In
  \bibinfo{booktitle}{\emph{2015 IEEE International Parallel and Distributed
  Processing Symposium}}. IEEE, \bibinfo{pages}{786--796}.
\newblock


\bibitem[\protect\citeauthoryear{et~al.}{et~al.}{2018i}]%
        {whatmough2018dnn}
\bibfield{author}{\bibinfo{person}{Whatmough~Paul et al.}}
  \bibinfo{year}{2018}\natexlab{i}.
\newblock \showarticletitle{DNN engine: A 28-nm timing-error tolerant sparse
  deep neural network processor for IoT applications}.
\newblock \bibinfo{journal}{\emph{IEEE Journal of Solid-State Circuits}}
  \bibinfo{volume}{53}, \bibinfo{number}{9} (\bibinfo{year}{2018}),
  \bibinfo{pages}{2722--2731}.
\newblock


\bibitem[\protect\citeauthoryear{et~al.}{et~al.}{2018j}]%
        {xygkis2018efficient}
\bibfield{author}{\bibinfo{person}{Xygkis~Athanasios et al.}}
  \bibinfo{year}{2018}\natexlab{j}.
\newblock \showarticletitle{Efficient winograd-based convolution kernel
  implementation on edge devices}. In \bibinfo{booktitle}{\emph{2018 55th
  ACM/ESDA/IEEE Design Automation Conference (DAC)}}. IEEE,
  \bibinfo{pages}{1--6}.
\newblock


\bibitem[\protect\citeauthoryear{et~al.}{et~al.}{2021d}]%
        {R2F2021TVLSI}
\bibfield{author}{\bibinfo{person}{Xu~Dawen et al.}}
  \bibinfo{year}{2021}\natexlab{d}.
\newblock \showarticletitle{R2F: A Remote Retraining Framework for AIoT
  Processors With Computing Errors}.
\newblock \bibinfo{journal}{\emph{IEEE Transactions on Very Large Scale
  Integration (VLSI) Systems}} \bibinfo{volume}{29}, \bibinfo{number}{11}
  (\bibinfo{year}{2021}), \bibinfo{pages}{1955--1966}.
\newblock
\urldef\tempurl%
\url{https://doi.org/10.1109/TVLSI.2021.3089224}
\showDOI{\tempurl}


\bibitem[\protect\citeauthoryear{et~al.}{et~al.}{2018k}]%
        {zhang2018thundervolt}
\bibfield{author}{\bibinfo{person}{Zhang~Jeff et al.}}
  \bibinfo{year}{2018}\natexlab{k}.
\newblock \showarticletitle{Thundervolt: enabling aggressive voltage
  underscaling and timing error resilience for energy efficient deep learning
  accelerators}. In \bibinfo{booktitle}{\emph{Proceedings of the 55th Annual
  Design Automation Conference}}. \bibinfo{pages}{1--6}.
\newblock


\bibitem[\protect\citeauthoryear{et~al.}{et~al.}{2017f}]%
        {zhao2017aep}
\bibfield{author}{\bibinfo{person}{Zhao~Lei et al.}}
  \bibinfo{year}{2017}\natexlab{f}.
\newblock \showarticletitle{AEP: An error-bearing neural network accelerator
  for energy efficiency and model protection}. In
  \bibinfo{booktitle}{\emph{2017 IEEE/ACM International Conference on
  Computer-Aided Design (ICCAD)}}. IEEE, \bibinfo{pages}{1047--1053}.
\newblock


\bibitem[\protect\citeauthoryear{Sparsh}{Sparsh}{2020}]%
        {mittal2020survey}
\bibfield{author}{\bibinfo{person}{Mittal Sparsh}.}
  \bibinfo{year}{2020}\natexlab{}.
\newblock \showarticletitle{A survey on modeling and improving reliability of
  DNN algorithms and accelerators}.
\newblock \bibinfo{journal}{\emph{Journal of Systems Architecture}}
  \bibinfo{volume}{104} (\bibinfo{year}{2020}), \bibinfo{pages}{101689}.
\newblock


\bibitem[\protect\citeauthoryear{Temam}{Temam}{2012}]%
        {temam2012defect}
\bibfield{author}{\bibinfo{person}{Olivier Temam}.}
  \bibinfo{year}{2012}\natexlab{}.
\newblock \showarticletitle{A defect-tolerant accelerator for emerging
  high-performance applications}. In \bibinfo{booktitle}{\emph{2012 39th Annual
  International Symposium on Computer Architecture (ISCA)}}. IEEE,
  \bibinfo{pages}{356--367}.
\newblock


\bibitem[\protect\citeauthoryear{Tu~Fengbin}{Tu~Fengbin}{2018}]%
        {tu2018rana}
\bibfield{author}{\bibinfo{person}{et~al. Tu~Fengbin}.}
  \bibinfo{year}{2018}\natexlab{}.
\newblock \showarticletitle{RANA: Towards efficient neural acceleration with
  refresh-optimized embedded DRAM}. In \bibinfo{booktitle}{\emph{2018 ACM/IEEE
  45th Annual International Symposium on Computer Architecture (ISCA)}}. IEEE,
  \bibinfo{pages}{340--352}.
\newblock


\end{thebibliography}




\end{document}